\documentclass[10pt,twocolumn,letterpaper]{article}

\usepackage{cvpr}
\usepackage{times}
\usepackage{epsfig}
\usepackage{graphicx}
\usepackage{amsmath}
\usepackage{amssymb}
\usepackage{booktabs}
\usepackage{subcaption}

\usepackage{multirow}
\usepackage{color}

\usepackage{authblk}

\usepackage[breaklinks=true,letterpaper=true,colorlinks,bookmarks=false]{hyperref}

 \cvprfinalcopy 

\def\cvprPaperID{***} 

 \ifcvprfinal\fi
\begin{document}

\title{Learning the Redundancy-free Features for Generalized Zero-Shot
  Object Recognition}

\newcommand*\samethanks[1][\value{footnote}]{\footnotemark[#1]}
\author[ ]{Zongyan Han\samethanks[2]}
\author[ ]{Zhenyong Fu\thanks{Corresponding authors.}\thanks{Zongyan
    Han, Zhenyong Fu and Jian Yang are with PCA Lab, Key Lab of
    Intelligent Perception and Systems for High-Dimensional
    Information of Ministry of Education, and Jiangsu Key Lab of Image
    and Video Understanding for Social Security, School of Computer
    Science and Engineering, Nanjing University of Science and
    Technology}}
\author[ ]{Jian Yang\samethanks[1]\samethanks[2]}


\affil[ ]{PCALab, Nanjing University of Science and Technology}
\affil[ ]{\tt\small {\{hanzy, z.fu, csjyang\}@njust.edu.cn}}

\maketitle
\thispagestyle{empty}

\begin{abstract}
Zero-shot object recognition or zero-shot learning aims to transfer
the object recognition ability among the semantically related
categories, such as fine-grained animal or bird species. However, 
the images of different fine-grained objects tend to merely exhibit
subtle differences in appearance, which will severely deteriorate
zero-shot object recognition. To reduce the superfluous information in
the fine-grained objects, in this paper, we propose to learn the
redundancy-free features for generalized zero-shot learning. We
achieve our motivation by projecting the original visual features into
a new (redundancy-free) feature space and then restricting the
statistical dependence between these two feature spaces. Furthermore,
we require the projected features to keep and even strengthen the
category relationship in the redundancy-free feature space. In this
way, we can remove the redundant information from the visual features
without losing the discriminative information. We extensively evaluate
the performance on four benchmark datasets. The results show that our
redundancy-free feature based generalized zero-shot learning
(RFF-GZSL) approach can achieve competitive results compared with the state-of-the-arts.
\end{abstract}

\vspace{-0.2cm}
\section{Introduction}
Object recognition has progressed remarkably in recent years thanks to
the deployments of deep Convolutional Neural Networks
(CNNs)~\cite{krizhevsky2012imagenet,he2016deep}. However, existing
CNN-based models, with tens to hundreds of millions of parameters,
excel only when large amounts of labeled data are available for each
object class, and generally struggle when labeled data are scarce. The
data-hungry nature of deep models limits their ability to recognize
rare object classes, such as fine-grained animal species. This is
because collecting and annotating a large set of images of these
classes is often labor-intensive and sometimes impossible (e.g.,
extinct species). Zero-Shot Learning (ZSL), also known as learning
from side information, provides a promising approach to addressing
this
problem~\cite{lampert2009learning,palatucci2009zero}. Specifically,
zero-shot learning aims to recognize the unseen classes, of which the
labeled images are unavailable, when the labeled images only from some
seen classes are provided~\cite{lampert2014attribute,xian2017zero}. In
ZSL, the seen classes are associated with the unseen classes in a
semantic descriptor space, such as the semantic attribute or word
vector space~\cite{akata2015evaluation}, which bridges the knowledge
gap between the seen and unseen classes.
  
\begin{figure}[t]
  \centering
  \begin{subfigure}[b]{0.115\textwidth}
    \centering
    \includegraphics[width=\textwidth]{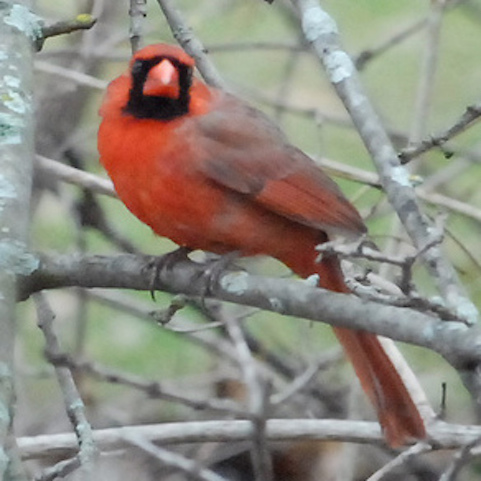}
    \caption{\footnotesize{Cardinal}}
    \label{fig:Cardinal}
  \end{subfigure}
  \begin{subfigure}[b]{0.115\textwidth}
    \centering
    \includegraphics[width=\textwidth]{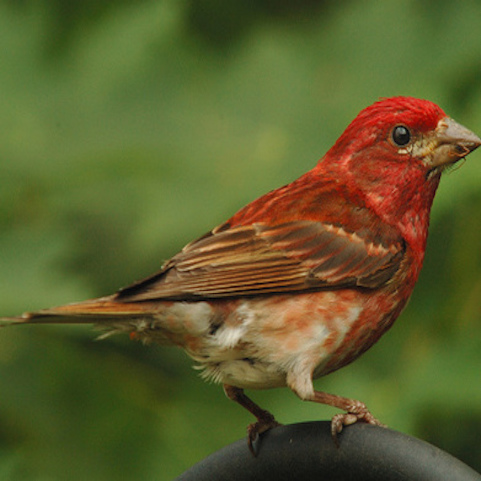}
    \caption{\footnotesize{Purple finch}}
    \label{fig:Purple_Finch}
  \end{subfigure}
  \begin{subfigure}[b]{0.115\textwidth}
    \centering
    \includegraphics[width=\textwidth]{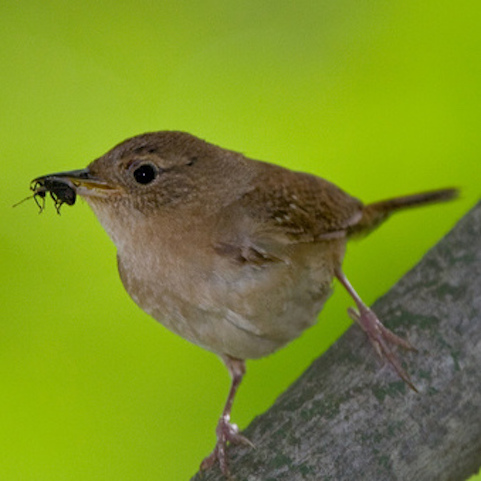}
    \caption{\footnotesize{House wren}}
    \label{fig:seal}
  \end{subfigure}
  \begin{subfigure}[b]{0.115\textwidth}
    \centering
    \includegraphics[width=\textwidth]{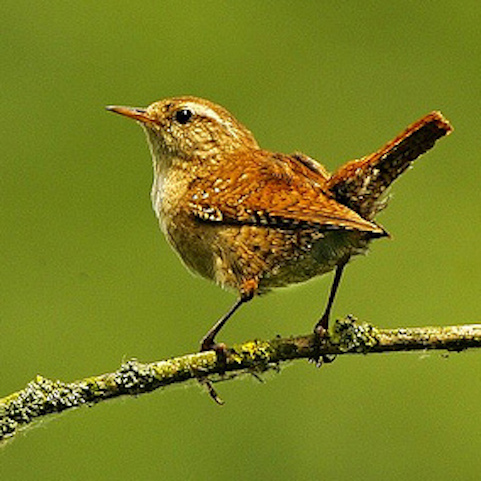}
    \caption{\footnotesize{Winter wren}}
    \label{fig:walrus}
  \end{subfigure}
  \caption{Examples of four fine-grained bird species. The subtle
    difference in appearance or the common living environment of these
    examples challenges the zero-shot object recognition.}
  \vspace{-0.2cm}
  \label{fig:animals}
\end{figure}

Although the conventional ZSL prevails in the early researches, the
realistic but more challenging Generalized Zero-Shot Learning (GZSL)
has drawn increasing attention recently. The conventional ZSL assumes
all test images coming from the unseen classes only, whereas the test
set in GZSL consists of data from both the seen and unseen classes.
Semantic embedding is the most important approach in conventional ZSL,
but generally performs poor in the new GZSL setting. The semantic
embedding
methods~\cite{frome2013devise,lampert2014attribute,akata2015evaluation}
learn to embed the visual features into the semantic descriptor space
and then predict the labels of visual features by finding their
nearest semantic descriptor. In GZSL, to mitigate the data imbalance
between seen and unseen classes, a number of feature generation
methods have been
proposed~\cite{xian2018feature,kumar2018generalized,felix2018multi,schonfeld2019generalized,xian2019f}.
The feature generation methods first learn a feature generator network
conditioned on the class-level semantic descriptors. The feature
generator can produce an arbitrary number of synthetic features and
thus compensate for the lack of visual features for the unseen
classes. In the end, the feature generation methods mix the real seen
features and the fake unseen features to train a supervised model,
e.g. a softmax classifier, as the final GZSL classifier.

Generalized zero-shot learning is usually evaluated on the
fine-grained datasets, such as Caltech-UCSD Birds
(CUB)~\cite{wah2011caltech}, as the fine-grained categories are
semantically related. The images of different fine-grained categories
tend to be very similar in appearance and merely exhibit subtle
differences, which will severely deteriorate the performance of the
GZSL classification. A similar background, such as a common living
environment, of fine-grained animal or bird species may also mislead
the zero-shot object recognition on these datasets, as shown in
Figure~\ref{fig:animals}. In other words, the fine-grained images in
GZSL contain superfluous content irrelevant to differentiating their
categories. Intuitively, GZSL can benefit from removing the redundancy
from the original visual features and preserving the most
discriminative information that triggers a class label.
  
In this paper, we present a generalized zero-shot learning approach
based on redundancy-free information. Specifically, we propose to map
the original visual features into a new space, where we bound the
dependence between the mapped features and the visual features to
remove the redundancy from the visual features. In the meanwhile, we
minimize the generalized ZSL classification error using the new
redundancy-free features to keep the discriminative information in
it. Our method is flexible in that it can be integrated with
the two aforementioned frameworks, i.e. semantic embedding and feature
generation, for GZSL. We evaluate our method on four widely used
datasets. The results show that when integrating with the conventional
semantic embedding framework, our model can surpass the other
conventional ZSL comparators in the new GZSL setting; when integrating
with the feature generation framework, to the best of our knowledge,
our model can achieve the state-of-the-art on two datasets and competitive results on the other two datasets.
This suggests that the
redundancy-free features are effective for the GZSL task.
  
Our contributions are three-fold: (1) we propose a redundancy-free
feature based GZSL method; (2) our method can integrate with the
conventional semantic embedding and the latest feature generation
frameworks; and (3) we evaluate our GZSL model on four benchmarks, and
to the best of our knowledge, our method can achieve the state-of-the-art or competitive results on these benchmarks.
  
\subsection{Related Work}
Zero-shot object recognition or zero-shot learning relies on the
class-level semantic descriptions or features, e.g. semantic
attributes~\cite{farhadi2009describing,parikh2011relative,akata2015evaluation}
and word vectors~\cite{mikolov2013efficient,mikolov2013distributed},
for model transferring
from the seen classes to a disjoint set of unseen classes. Earlier ZSL
research works focus on the conventional ZSL
problem~\cite{norouzi2013zero,socher2013zero,frome2013devise,akata2015evaluation,
  zhang2015zero,fu2015zero,kodirov2015unsupervised,bucher2016improving,
  xian2016latent,fu2017zero,changpinyo2016synthesized,romera2015embarrassingly,
  kodirov2017semantic}, in which the semantic embedding is the most
important approach~\cite{frome2013devise,romera2015embarrassingly,
  bucher2016improving,kodirov2017semantic}. Semantic embedding methods
learn to embed the visual features into the semantic descriptor space,
or vice
versa~\cite{frome2013devise,akata2013label,akata2015evaluation,kodirov2017semantic}. By
doing so, the visual features and the semantic features will lie in a
same space and the ZSL classification can be accomplished by searching
the nearest semantic descriptor.
  
In the more challenging GZSL task, we have the labeled data only from
seen classes during training, but need to recognize the images from
both seen and unseen classes in the test phase. Thus, GZSL suffers
from the extreme data imbalance problem. Semantic embedding methods
fail to solve the data imbalance problem in GZSL. They tend to be
highly overfitting the seen classes and thus harm the classification
of unseen classes. The experiments in~\cite{xian2017zero} showed that
the performance of almost all conventional ZSL methods, including
semantic embedding, drops significantly in the new GZSL scenario.
  
To compensate for the lack of training images of unseen classes in
GZSL, recently, some feature generation methods have been
proposed to tackle the GZSL problem~\cite{bucher2017generating,
  xian2018feature,felix2018multi,kumar2018generalized,xian2019f,
  huang2019generative,schonfeld2019generalized}. Bucher et
al.~\cite{bucher2017generating} proposed to generate features for
unseen classes with four different generative models, including
generative moment matching network (GMMN)~\cite{li2015generative},
auxiliary classifier GANs (AC-GAN)~\cite{odena2017conditional},
denoising auto-encoder~\cite{bengio2013generalized} and adversarial
auto-encoder (AAE)~\cite{makhzani2015adversarial}. The f-CLSWGAN
in~\cite{xian2018feature} proposed to generate the unseen features
conditioned on the class-level semantic descriptors. Some
methods~\cite{felix2018multi,huang2019generative} further constrained
the feature generator network by introducing a reverse regressor
network which can be used to define a cycle-consistent
loss~\cite{zhu2017unpaired}. Verma et al.~\cite{kumar2018generalized} 
built their feature synthesis framework upon Variational Autoencoder
(VAE)~\cite{kingma2013auto}. Besides the feature generation methods,
Chen et al.~\cite{chen2018zero} proposed an adversarial
visual-semantic embedding framework. Liu et
al.~\cite{liu2018generalized} proposed a deep calibration network
(DCN) that simultaneously calibrates the model confidence on seen
classes and the model uncertainty on unseen classes.
  
As previously analyzed, the images of different fine-grained
categories in GZSL differ slightly in appearance, which will challenge  
the GZSL classification. To mitigate this problem, we propose to
reduce the redundant information in the visual features for GZSL. Our
work is inspired by the information bottleneck
method~\cite{alemi2016deep}. Concretely, we map the visual features
into a new redundancy-free feature space and limit the information
dependence between the mapped features and the original images
features to an upper bound. 

\section{Preliminaries}
In this section, we define the GZSL problem and then briefly revisit
the semantic embedding and feature generation frameworks in GZSL.

\vspace{-0.2cm}
\paragraph{Problem definition}
In zero-shot learning, we are given a set of seen classes
$\mathcal{Y}_s$ and a disjoint set of unseen classes $\mathcal{Y}_u$,
where we have $\mathcal{Y}_s\cap\mathcal{Y}_u=\varnothing$. Suppose
that there is a training dataset
$\mathcal{D}^{tr}_s=\{(x_i,a_i,y_i)\}$ consisting of labeled samples
from the \emph{seen} classes only, where $x_i\in X$ represents the
visual feature, $a_i\in\mathcal{A}$ is the associated semantic
descriptor (e.g. semantic attributes), and $y_i\in\mathcal{Y}_s$
denotes the seen class label. The semantic features of unseen classes
are also available, but their visual features are missing. Zero-shot
learning aims to learn a classifier being evaluated on a test dataset
$\mathcal{D}^{te}=\{x_k\}$. In generalized ZSL, the test dataset
$\mathcal{D}^{te}$ is composed of examples from both seen and unseen
classes, i.e., GZSL is tested on $\mathcal{Y}_s\cup\mathcal{Y}_u$.

\vspace{-0.2cm}
\paragraph{Semantic embedding}
The conventional semantic embedding methods in ZSL learn an embedding
function $E$ that maps a visual feature $x$ into the semantic
descriptor space as $E(x)$. In this paper, we adopt a structured
objective proposed in~\cite{akata2015evaluation,frome2013devise} to
learn the embedding function $E$. Such a structured objective requires
the embedding of $x$ being closer to the semantic descriptor $a$ of
its ground-truth class than the descriptors of other classes,
according to the dot-product similarity in semantic descriptor
space. This objective for learning $E$ is defined as below:
\begin{equation}
  \min_{E}\mathbb{E}_{p(x,a)}[\max(0,\Delta-a^{\top}E(x)+(a^\prime)^{\top}E(x))],
  \label{eq:SJE}
\end{equation}where $p(x,a)$ is the empirical data distribution of
seen classes defined on $\mathcal{D}^{tr}_s$, $a^\prime\neq a$ is a
randomly-selected semantic descriptor of other classes, and $\Delta>0$
is a margin to make $E$ more robust. Once the embedding function $E$
is optimized, we can use $E$ to embed the visual feature of a test
image to the semantic descriptor space and infer its class label by
finding the nearest semantic descriptor.

\vspace{-0.2cm}
\paragraph{Feature generation}
Semantic embedding methods prevail in the conventional ZSL but is
unsuccessful in the more challenging generalized ZSL problem. Feature
generation can address the data imbalance problem in GZSL and their
effectiveness for GZSL has been evidenced recently~\cite{xian2018feature,kumar2018generalized,mishra2018generative,huang2019generative,atzmon2019adaptive}. We
adopt a basic feature generation method, f-CLSWGAN, proposed
in~\cite{xian2018feature}, although our approach based on
redundancy-free features can certainly integrate with other more 
sophisticated feature generation methods. f-CLSWGAN learns a visual
feature generator $G$, defined as a conditional generative model
$\tilde{x}=G(a,\epsilon)$, conditioned on a semantic descriptor $a$
and a Gaussian noise
$\epsilon\sim\mathcal{N}(\mathbf{0},\mathbf{I})$. In f-CLSWGAN, a
discriminator $D$ is learned together with $G$ to discriminate a real 
pair $(x,a)$ from a synthetic pair $(\tilde{x},a)$, whereas the
feature generator $G$ tries to fool the discriminator $D$ by producing
indistinguishable synthetic features. As shown
in~\cite{goodfellow2014generative}, such an idea can be formulated as
the following adversarial objective:
\begin{equation}
  \min_{G}\max_{D}\mathbb{E}_{p(x,a)}[\log D(x,a)]
  +\mathbb{E}_{p_{G}(\tilde{x})}[\log(1-D(\tilde{x},a))],
  \label{eq:loss_adv}
\end{equation}where $p_{G}$ is the distribution of synthetic visual
features. To make the generated visual features more discriminative,
f-CLSWGAN further constrains the generator $G$ with a supervised
classification loss:
\begin{equation}
  \mathcal{L}_{CLS}(G)=-\mathbb{E}_{p_{G}(\tilde{x})}[\log q(y|\tilde{x})],
  \label{eq:loss_cls}
\end{equation}where $q(y|\tilde{x})$ is a classifier that is
pre-trained on the seen training set
$\mathcal{D}^{tr}_s$. $q(y|\tilde{x})$ gives the probability of
$\tilde{x}$ being predicted as the label $y$ inherited from the
conditional semantic descriptor $a$. The feature generator $G$ can
synthesize an arbitrary number of labeled features for unseen
classes. As a result, we can transform GZSL to a standard supervised
learning problem.
  
\section{Methodology}
In this section, we present how to learn the redundancy-free
information and then describe how it can be integrated with the
semantic embedding and the feature generation frameworks,
respectively, to tackle the GZSL problem.
  
\subsection{Learning the Redundancy-free Information}
We learn a mapping function $M$ to map the original visual features to
a new feature space. Our goal is to remove the redundancy information
contained in the original feature $x$ through $M$; $z=M(x)$ represents
the redundancy-free information of $x$. Let $X$ be the original
features and $Z$ be the redundancy-free features. We hope to perform
the GZSL task using the redundancy-free features rather than the
original redundancy features. To the end, we bound the 
statistical dependence between $Z$ and $X$ to enforce $Z$ to forget
the redundancy information in $X$. In information theory, the
dependence between two random variables is measured by mutual
information (MI) $I(Z;X)$, defined as $I(Z;X)=H(Z)-H(Z|X)$, where
$H(Z)$ is the marginal entropy of $Z$ and $H(Z|X)$ is the conditional
entropy of $Z$ with respect to $X$. Note that we do not intend to
minimize the mutual information $I(Z;X)$ but ask it to be lower than
an upper bound, such that some information in $X$ can still be
conveyed to $Z$. Otherwise, $I(Z;X)=0$ means $Z$ and $X$ are
statistically independent.
  
Calculating the mutual information with high dimension is
intractable. We follow the strategy proposed by Alemi et
al.~\cite{alemi2016deep} to use a variational upper bound of MI as a
surrogate:
\begin{equation}
  I(Z;X)\leq\mathbb{E}_{p(x)}[D_{KL}[p_M(z|x)\|r(z)]],
  \label{eq:variational_mi}
\end{equation}where $D_{KL}()$ is the Kullback-Leibler (KL)
divergence, $p_{M}(z|x)$ is the conditional distribution of
redundancy-free features $z$ conditioned on the original visual
features $x$. $r(z)$ is the variational approximation to the marginal
distribution of $z$. The variational upper bound can be estimated
using the reparameterization trick~\cite{kingma2013auto}. By
restricting this variational upper bound, we can implicitly constrain
the mutual information between $Z$ and $X$. In this way, the mapping
function $M$ can be learned to extract the redundancy-free information
from $x$.

Only removing the redundancy information from the original visual
features cannot guarantee a satisfactory GZSL result. Next, we will
discuss how to preserve the discriminative information concerning
GZSL, in $z$.
  
\subsection{Redundancy-free Semantic Embedding for GZSL}
To exploit the redundancy-free information in the semantic embedding
methods, we simply regard the semantic descriptor space as the new
feature space and request the function $M$ to map the original
visual features into the semantic descriptor space, analogously to the
conventional semantic embedding method described above. Therefore, we
just constrain the structured objective defined in Eq.~\ref{eq:SJE}
with the bounded variational mutual information as below:
\begin{equation}
  \begin{aligned} 
    \min_{M}&\hspace{0.5em}\mathbb{E}_{p(x,a)}[\mathbb{E}_{p_{M}(z|x)}[\max(0,\Delta-a^{\top}z+(a^{\prime})^{\top}z)]]\\
    \text{s.t.}&\hspace{0.5em}\mathbb{E}_{p(x)}[D_{KL}[p_{M}(z|x) \| r(z)]]\leq b,
  \end{aligned}
  \label{eq:our_embedding_st}
\end{equation}where $b$ is the upper bound we impose. We apply the
strategy described in~\cite{peng2018variational} to optimize an
unconstrained form of Eq.~\ref{eq:our_embedding_st} derived by the
method of Lagrange multiplier.

\begin{figure}[t]
  \centering
  \includegraphics[width=\linewidth]{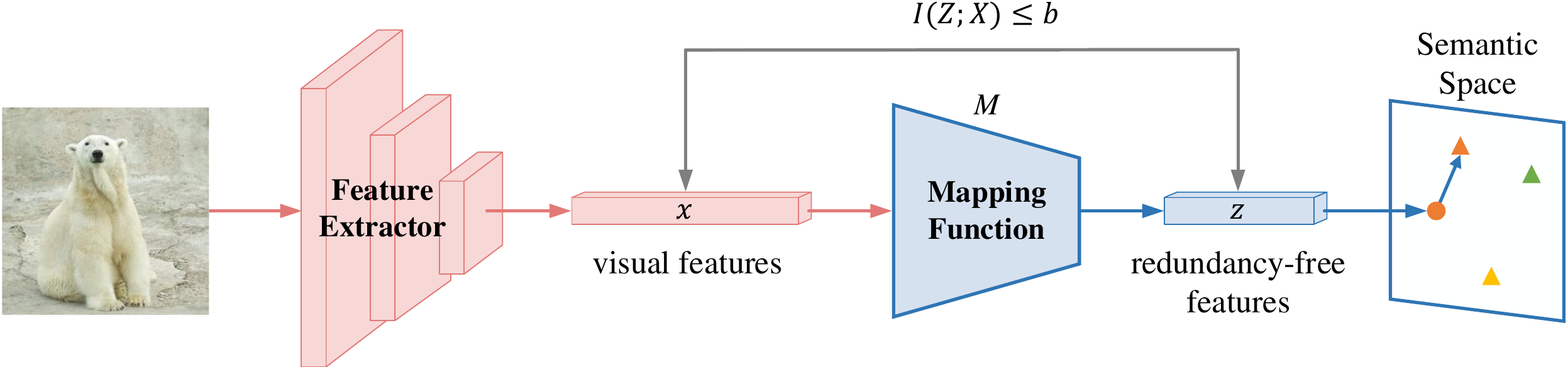}
  \caption{The structure of the redundancy-free \emph{semantic embedding}
    framework for GZSL. We learn a mapping function $M$
    to project a visual feature to the semantic descriptor space. We
    bound the statistical dependence measured by the mutual
    information between the mapped features and the original visual
    features to enforce $M$ to extract the redundancy-free information
    from the visual features.}
  \vspace{-0.2cm}
  \label{fig:embedding}
\end{figure}

Figure~\ref{fig:embedding} shows the schematic overview of the
redundancy-free semantic embedding method. Our method differs from the
traditional semantic embedding methods in that we restrict the
embedding $z=M(x)$ to preserve the information in the original feature
$x$ to an upper bound. More specifically, in
Eq.~\ref{eq:our_embedding_st}, the information constraint determines
how much information in $x$ will be conveyed to $z$, and the
classification term decides whether the left information is
discriminative for GZSL or not.
  
\begin{figure*}[t]
  \centering
  \includegraphics[width=\linewidth]{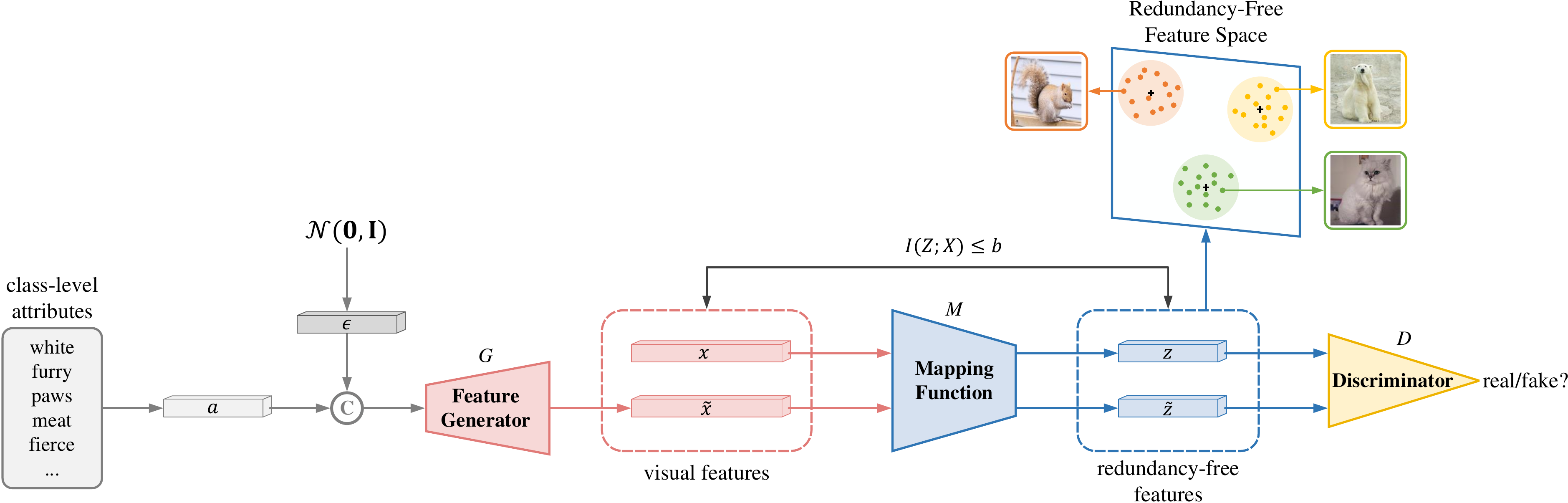}
  \caption{The structure of the redundancy-free \emph{feature generation}
    framework for GZSL. We learn a feature generator $G$ that
    synthesizes the fake visual features for unseen
    classes using the class-level semantic descriptor $a$
    (e.g. attributes) and a noise $\epsilon$. Furthermore, we learn a
    mapping function $M$ to map the real visual features of seen
    classes and the synthetic visual features of unseen classes into a
    new redundancy-free feature space. We remove the redundancy
    information from the visual features by restricting the mutual
    information between the original visual features and the
    redundancy-free features. Our redundancy-free features are 
    discriminative for the GZSL task.}
  \vspace{-0.2cm}
  \label{fig:generation}
\end{figure*}
  
\subsection{Redundancy-free Feature Generation for GZSL}
Previous feature generation methods in GZSL trained a feature generator
network to mimic the distribution of real visual features. To exploit
the redundancy-free information in feature generation, we take one
step further and learn a new mapping function $M$ to project the
visual features, real and synthetic, to another feature space.
For the seen classes in GZSL, we use the new mapping function $M$ to
transform the original real visual features to the redundancy-free
features: $z=M(x)$. For the unseen classes in GZSL, we indeed use a
composite generator network $M\circ G$ to synthesize the fake and
redundancy-free features: $\tilde{z}=(M\circ
G)(a,\epsilon)=M(G(a,\epsilon))$. In this way, we can rewrite
the adversarial objective of feature generation defined in
Eq.~\ref{eq:loss_adv} as follows:
\begin{equation}
  \begin{aligned}
    V(D,M\circ G)&=
    \mathbb{E}_{p(x)}[\mathbb{E}_{p_{M}(z|x)}[\log D(z)]]\\
    &\hspace{0.1em}+\mathbb{E}_{p_{G}(\tilde{x})}[\mathbb{E}_{p_{M}(\tilde{z}|\tilde{x})}[\log(1-D(\tilde{z}))]],
  \end{aligned}
\end{equation}where $p_{M}(\tilde{z}|\tilde{x})$ is the distribution
of the synthesized redundancy-free features $\tilde{z}$ conditioned on
the synthetic visual features $\tilde{x}$.

To ensure the generated features have a similar discriminative ability
like the real feature, f-CLSWGAN further constrained the feature
generator network $G$ with a pre-trained supervised classifier given
in Eq.~\ref{eq:loss_cls}. Similarly, to ensure the redundancy-free
features produced by $M$ are also discriminative, we use the training
visual features of seen classes to constrain $M$ so that the category
relationship of the seen training data can be well retained in the
redundancy-free feature space. Concretely, we constrain the mapping
function $M$ using the following loss objective:
\begin{equation}
  \mathcal{L}_r(M,\mathbf{c})=\mathbb{E}_{p(x,y)}[\mathbb{E}_{p_{M}(z|x)}[\mathcal{L}_c(z,y,y^\prime)]],
  \label{eq:loss_center}
\end{equation}in which for each sample from seen classes, we
compute the loss of its redundancy-free feature with the center loss
proposed in~\cite{wen2016discriminative} as below:
\begin{equation}
  \mathcal{L}_c(z, y,
  y^\prime)=\max(0,\Delta+\|z-\mathbf{c}_y\|^2_2-\|z-\mathbf{c}_{y^{\prime}}\|^2_2).
  \label{eq:loss_c}
\end{equation}where $y$ is the class label of $x$ and $y^{\prime}$ is
a randomly-selected class label other than $y$. In
$\mathcal{L}_r(M,\mathbf{c})$, an array of centers in the
redundancy-free feature space, one for each seen class, are optimized 
with $M$ together. The center loss can group the redundancy-free
features of seen classes according to their labels such that the
distributions of different classes can be easily separated. As such,
we indeed strengthen the category relationships of seen class data in
the new redundancy-free feature space.

We formulate our final learning objective for redundancy-free feature
generation as follows:
\begin{small}
  \begin{equation}
    \begin{aligned}
      \min
      _{G,M,\mathbf{c}}\max_{D}&\hspace{0.3em}V(D,M\circ G)
      +\lambda_r\mathcal{L}_r(M,\mathbf{c})+\lambda_c\mathcal{L}_{CLS}(G)\\
      \text{s.t.}&\hspace{0.3em}\mathbb{E}_{p(x)}[D_{KL}[p_{M}(z|x) \|
      r(z)]]\leq b\\
      &\hspace{0.3em}\mathbb{E}_{p_G(\tilde{x})}[D_{KL}[p_{M}(\tilde{z}|\tilde{x}) \|
      r(\tilde{z})]]\leq b.
    \end{aligned}
    \label{eq:loss_all}
  \end{equation}
\end{small}In Eq.~\ref{eq:loss_all}, we learn the discriminator $D$
and the composite generator $M\circ G$ in an adversarial manner, to
avoid the mismatching between the distribution of synthetic
redundancy-free features and that of real redundancy-free
features. We keep the classification loss $\mathcal{L}_{CLS}$
(Eq.~\ref{eq:loss_cls}) in the original visual feature space, to
ensure the discriminative ability of the generated unseen visual
features, which will be mapped to the new redundancy-free feature
space later. The two information constraints bound the variational
mutual information so that the redundancy information can be removed
from the visual features. Last, $\mathcal{L}_r$ will encourage $M$ to
produce the well-separated thus strongly discriminative
redundancy-free features. Figure~\ref{fig:generation} shows the
overall structure of the redundancy-free feature generation
framework.
  
\subsection{Classification}
\paragraph{Semantic embedding}
For a test data point $x\in\mathcal{D}^{te}$, we use $M$ to map it
into the semantic descriptor space as $M(x)$. $x$ will be labeled as
the class with the nearest semantic descriptor with respect to $x$:
\begin{equation}
  y^{*}=\arg\max_{a\in\mathcal{A}_s\cup\mathcal{A}_u}a^{\top}M(x).
  \label{eq:cls_embed}
\end{equation}

\vspace{-0.3cm}
\paragraph{Feature Generation}
We first map all training data of seen classes into the
redundancy-free feature space as $z=M(x)$ for each $x\in D^{tr}_s$. We
then synthesize a set of redundancy-free features for each unseen 
class $y\in\mathcal{Y}_u$ by performing $\tilde{z}=(M\circ
G)(a_y,\epsilon)$. Once we have the training data, real or fake for
each seen or unseen class, we train a supervised classifier in the
redundancy-free feature space as the final GZSL classifier. In this
paper, we evaluate our method with softmax classifiers.

\begin{table*}[t]
  \caption{Results of the state-of-the-arts. $U$ and $S$ are the Top-1
    accuracies tested on unseen classes and seen classes,
    respectively, in GZSL. $H$ is the harmonic mean of $U$ and $S$. On each dataset, we synthesize different numbers
    of examples per unseen class: AWA (1800), CUB (400), SUN
    (400), and FLO (1200). \textdaggerdbl\hspace{0.05em} and
    \textdagger\hspace{0.05em} denote the feature generation methods
    or not, respectively. The best results
and the second best results are respectively marked in bold and underlined.}
  \vspace{-0.2cm}
  \centering
  \resizebox{0.85\textwidth}{!}
  {
  \label{table:comp_sota}
  \begin{tabular}{c|l|ccc|ccc|ccc|ccc}
    \toprule
    &\multirow{2}*{Method} &\multicolumn{3}{|c|}{AWA}&\multicolumn{3}{c|}{CUB}&\multicolumn{3}{c|}{SUN}&\multicolumn{3}{c}{FLO}\\
    &&${U}$&${S}$&${H}$&${U}$&${S}$&${H}$&${U}$&${S}$&${H}$&${U}$&${S}$&${H}$\\
    \midrule
    \multirow{5}*{
    \begin{tabular}{c}
      \textdagger
    \end{tabular}
   }
    &DCN~\cite{liu2018generalized}&25.5&\underline{84.2}&39.1&28.4&60.7&38.7&25.5&37.0&30.2&-&-&-\\
    &SP-AEN~\cite{chen2018zero}&23.3&\textbf{90.9}&37.1&34.7&\underline{70.6}&46.6&24.9&38.6&30.3&-&-&-\\
    &AREN~\cite{xie2019attentive}&-&-&-&38.9&\textbf{78.7}&52.1&19.0&\underline{38.8}&25.5&-&-&-\\
    &Kai at el.~\cite{li2019rethinking}&\textbf{62.7}&77.0&\textbf{69.1}&47.4&47.6&47.5&36.3&\textbf{42.8}&39.3&-&-&-\\
    &CRnet~\cite{zhang2019co}&58.1&74.7&65.4&45.5&56.8&50.5&34.1&36.5&35.3&-&-&-\\
    \midrule
    \multirow{9}*{
      \begin{tabular}{c}
        \textdaggerdbl
      \end{tabular}
    }
    &SE-GZSL~\cite{kumar2018generalized}&56.3&67.8&61.5&41.5&53.3&46.7&40.9&30.5&34.9&-&-&-\\
    &f-CLSWGAN~\cite{xian2018feature}&57.9&61.4&59.6&43.7&57.7&49.7&42.6&36.6&39.4&59.0&73.8&\underline{65.6}\\
    &cycle-CLSWGAN~\cite{felix2018multi}&56.9&64.0&60.2&45.7&61.0&52.3&49.4&33.6&40.0&\underline{59.2}&72.5&65.1\\
    &CADA-VAE~\cite{schonfeld2019generalized}&57.3&72.8&64.1&51.6&53.5&52.4&47.2&35.7&40.6&-&-&-\\
    &SABR~\cite{paul2019semantically}&-&-&-&\underline{55.0}&58.7&\textbf{56.8}&\underline{50.7}&35.1&\underline{41.5}&-&-&-\\    
    &f-VAEGAN~\cite{xian2019f}&-&-&-&48.4&60.1&53.6&45.1&38.0&41.3&56.8&74.9&64.6\\
    &LisGAN~\cite{li2019leveraging}&52.6&76.3&62.3&46.5&57.9&51.6&42.9&37.8&40.2&57.7&\textbf{83.8}&68.3 \\
    &GMN~\cite{sariyildiz2019gradient}&\underline{61.1}&71.3&65.8&\textbf{56.1}&54.3&\underline{55.2}&\textbf{53.2}&33.0&40.7&-&-&-\\
    \cline{2-14}
    &\textbf{Our RFF-GZSL} &59.8&75.1&\underline{66.5}&52.6&56.6&54.6&45.7&38.6&\textbf{41.9}&\textbf{65.2}&\underline{78.2}&\textbf{71.1}\\
    \bottomrule
  \end{tabular}
  }
  \vspace{-0.2cm}
\end{table*}

\section{Experiments}
\paragraph{Datasets}
We evaluate our method on four datasets for GZSL: (1) Animals with
Attributes 1 (\textbf{AWA})~\cite{lampert2009learning} consists of 50
classes of animals with 30,475 examples annotated with 85 attributes;
(2) Caltech-UCSD Birds-200-2011 (\textbf{CUB})~\cite{wah2011caltech}
contains 11,788 examples of 200 fine-grained bird species annotated
with 312 attributes;
(3) SUN Attribute (\textbf{SUN})~\cite{patterson2012sun} consists of
14,340 examples of 717 different scenes annotated with 102 attributes;
(4) Oxford Flowers (\textbf{FLO})~\cite{nilsback2008automated} is
composed of 8,189 examples of 102 different fine-grained flower
classes annotated with 1,024 attributes~\cite{reed2016learning}.
We extract the 2,048-dimensional CNN features for images using
ResNet-101~\cite{he2016deep} as the visual features and the
pre-defined attributes on each dataset are used as the semantic
descriptors. Moreover, we adopt the Proposed Split
(PS)~\cite{xian2017zero} to divide the total classes into seen and
unseen classes on each dataset.

\vspace{-0.2cm}
\paragraph{Evaluation Protocols}
The performances of our method are evaluated by per-class Top-1
accuracy. In GZSL, since the test set is composed of seen and unseen
images, we will evaluate the Top-1 accuracies respectively on seen
classes, denoted as $S$, and unseen classes, denoted as $U$. Their
harmonic mean, defined as $H=(2\times S\times U)/
(S+U)$~\cite{xian2017zero}, evaluates the performance of GZSL.

\vspace{-0.2cm}
\paragraph{Implementation Details}
We implement our model with neural networks using PyTorch. The
generator $G$ contains a 4096-unit hidden layer with LeakyReLU
activation. The mapping function $M$ and discriminator $D$ is
implemented with a fully-connected layer and ReLU activation. We use
Adam solver~\cite{kingma2014adam} with $\beta_1=0.5, \beta_2=0.999$
and a batch size of 512.
We empirically set the MI bound $b=0.1$, the dimension of
redundancy-free feature space as 1,024 and $\lambda_r=0.1$; we
cross-validate $\lambda_c$ in $[0.1, 1]$. To make the training process
more stable, we adopt Wasserstein GAN~\cite{arjovsky2017wasserstein}
and some improved strategies~\cite{gulrajani2017improved} in the
feature generation framework.

\begin{figure}[t]
  \centering
  \begin{subfigure}[b]{0.23\textwidth}
    \includegraphics[width=\linewidth]{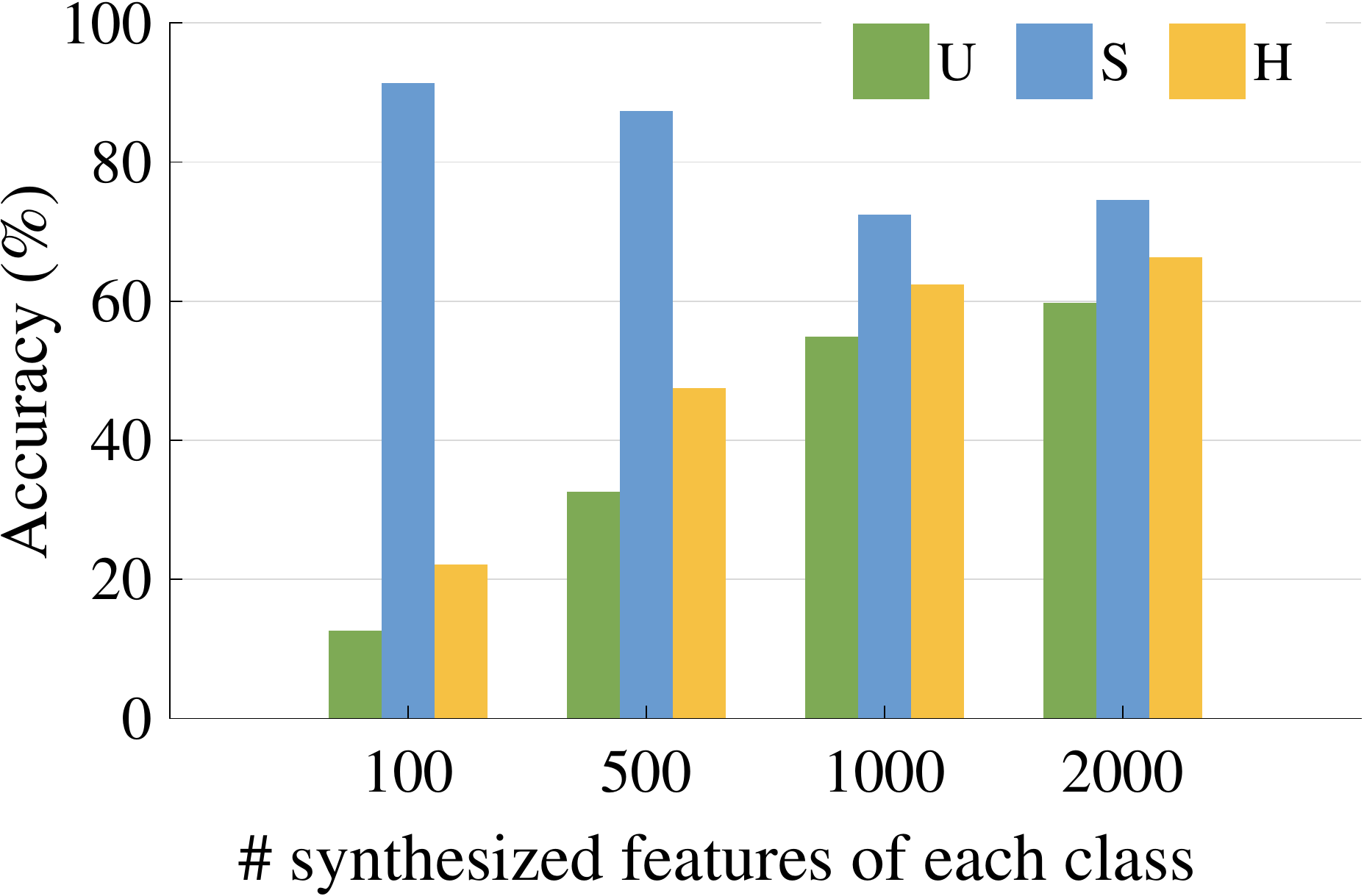}
    \caption{AWA}
    \label{fig:syn_num_AWA}
  \end{subfigure}\hspace{0.006\textwidth}
  \begin{subfigure}[b]{0.23\textwidth}
    \includegraphics[width=\linewidth]{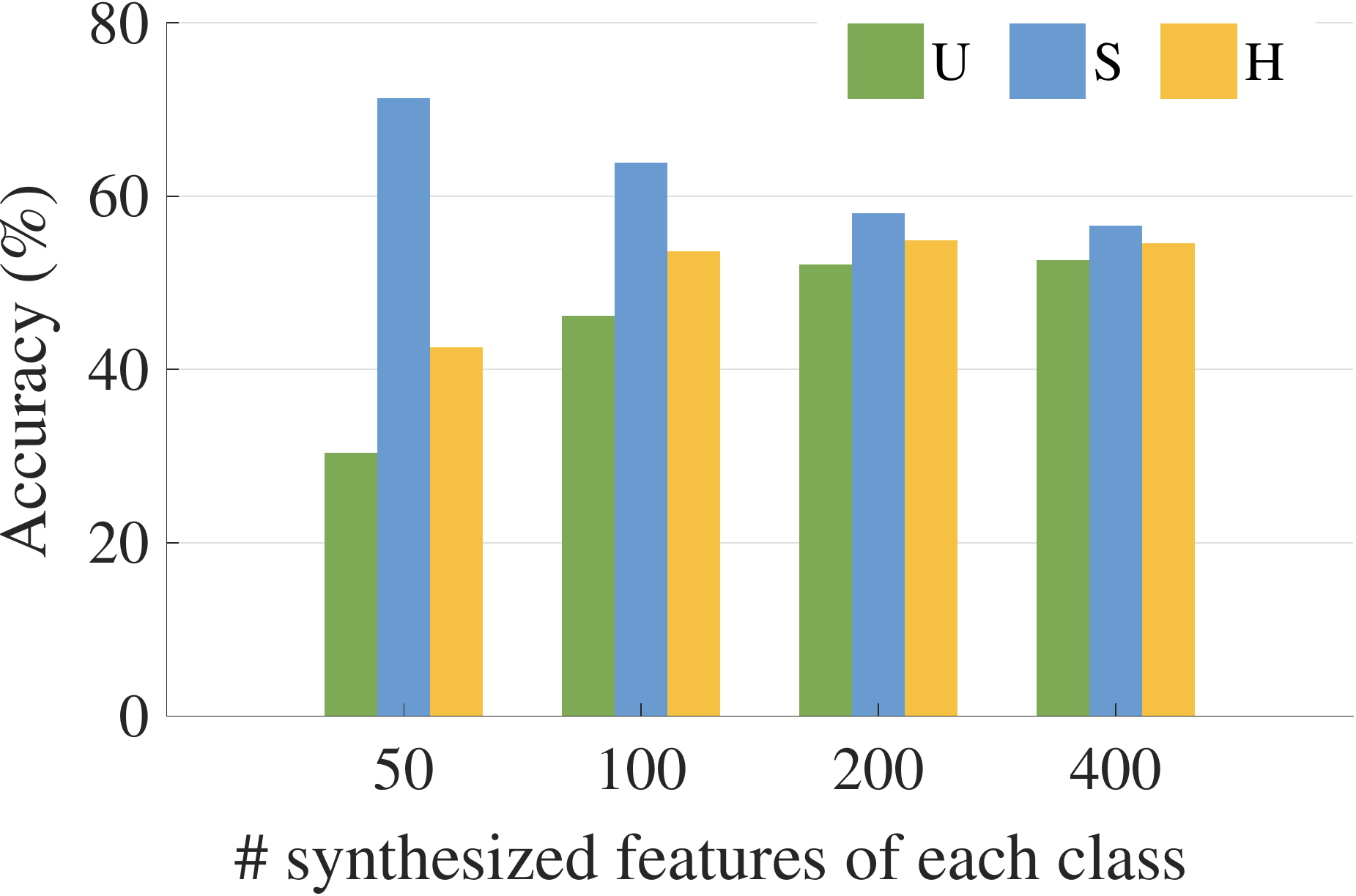}
    \caption{CUB}
    \label{fig:syn_num_CUB}
  \end{subfigure}\hspace{0.006\textwidth}
  \begin{subfigure}[b]{0.23\textwidth}
    \includegraphics[width=\linewidth]{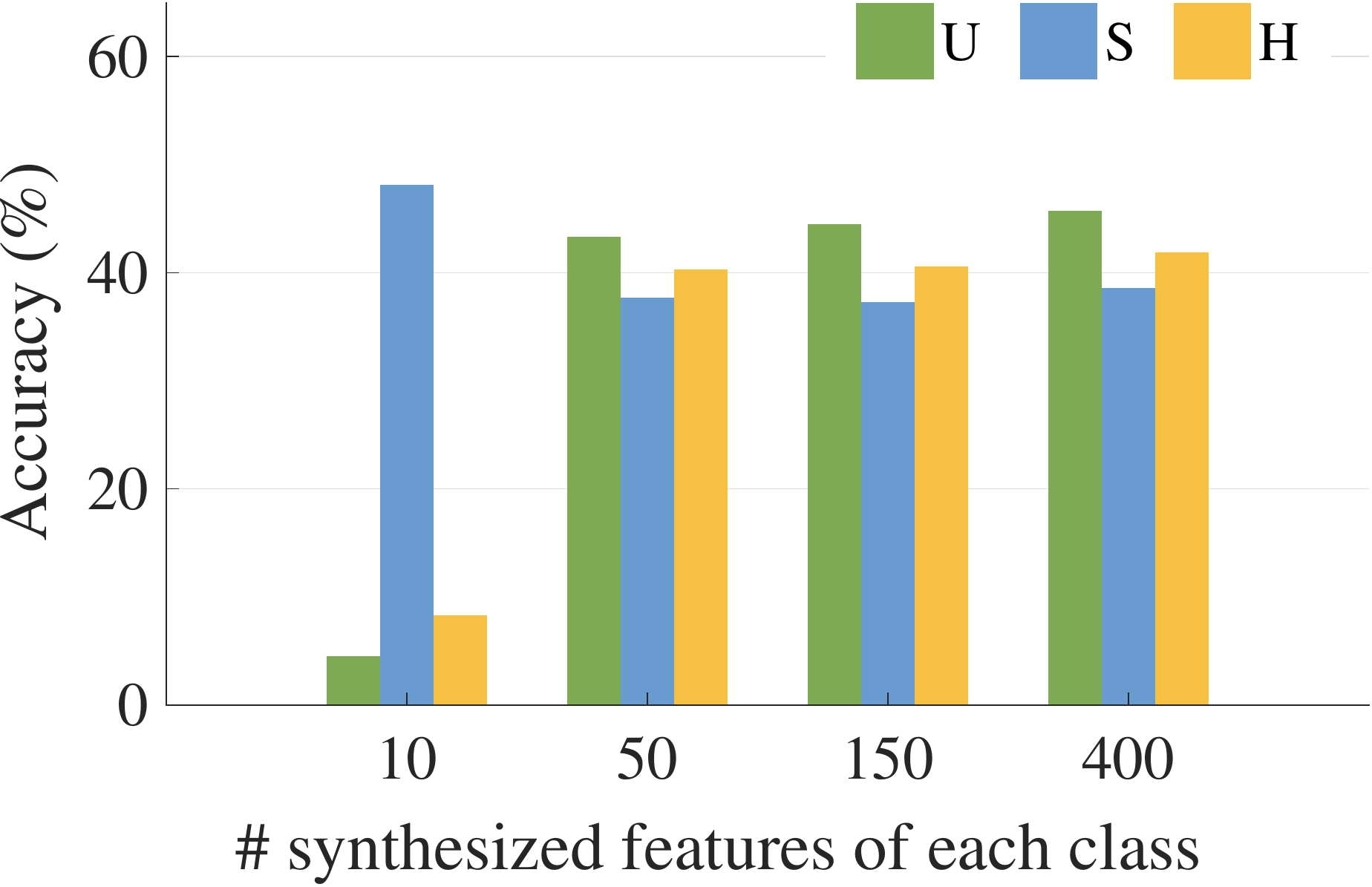}
    \caption{SUN}
    \label{fig:syn_num_SUN}
  \end{subfigure}\hspace{0.006\textwidth}
  \begin{subfigure}[b]{0.23\textwidth}
    \includegraphics[width=\linewidth]{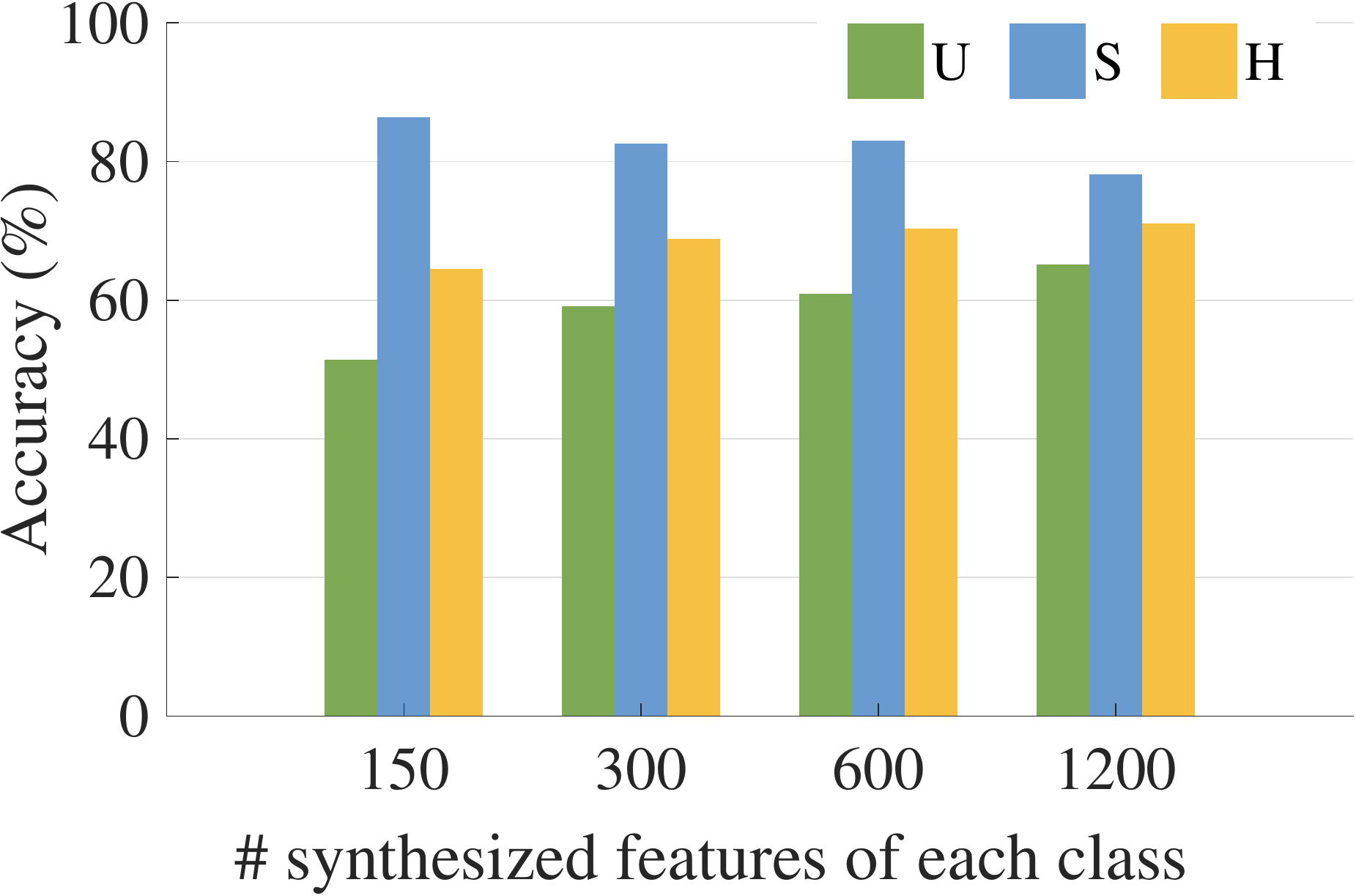}
    \caption{FLO}
    \label{fig:syn_num_FLO}
  \end{subfigure}
  \caption{The GZSL results of our RFF-GZSL with respect to
    different numbers of synthetic samples per unseen class.}
  \vspace{-0.2cm}
  \label{fig:syn_num}
\end{figure}

\begin{table*}[ht]
  \caption{Results of comparison with traditional ZSL methods in the
    new GZSL scenario. $U$ and $S$ are the Top-1 accuracies tested on
    unseen classes and seen classes, respectively, in GZSL. $H$ is the
    harmonic mean of $U$ and $S$.
    The best results
    and the second best results are respectively marked in bold and underlined.}
  \centering
  \resizebox{0.76\textwidth}{!}
  {
  \label{tab:em_gzsl}
  \begin{tabular}{l|ccc|ccc|ccc|ccc}
    \toprule
    \multirow{2}*{Method} &\multicolumn{3}{|c|}{AWA}&\multicolumn{3}{c|}{CUB}&\multicolumn{3}{c|}{SUN}&\multicolumn{3}{c}{FLO}\\
                          &${U}$&${S}$&${H}$&${U}$&${S}$&${H}$&${U}$&${S}$&${H}$&${U}$&${S}$&${H}$\\
    \midrule
    DAP~\cite{lampert2014attribute}&0.0&\textbf{88.7}&0.0&1.7&67.9&3.3&4.2&25.1&7.2&-&-&-\\
    IAP~\cite{lampert2014attribute}&2.1&78.2&4.1&0.2&\textbf{72.8}&0.4&1.0&\underline{37.8}&1.8&-&-&-\\
    SJE~\cite{akata2015evaluation}&11.3&74.6&19.6&23.5&59.2&33.6&14.7&30.5&19.8&\underline{13.9}&47.6&21.5\\
    LATEM~\cite{xian2016latent}&7.3&71.7&13.3&15.2&57.3&24.0&14.7&28.8&19.5&6.6&47.6&11.5\\
    DEVISE~\cite{frome2013devise}&13.4&68.7&22.4&\underline{23.8}&53.0&32.8&16.9&27.4&20.9&9.9&44.2&16.2\\
    ALE~\cite{akata2013label}&\underline{16.8}&76.1&\underline{27.5}&23.7&62.8&\underline{34.4}&\underline{21.8}&33.1&\underline{26.3}&13.3&\underline{61.6}&\underline{21.9}\\
    ESZSL~\cite{romera2015embarrassingly}&6.6&75.6&12.1&12.6&63.8&21.0&11.0&27.9&15.8&11.4&56.8&19.0\\
    SYNC~\cite{changpinyo2016synthesized}&8.9&\underline{87.3}&16.2&11.5&\underline{70.9}&19.8&7.9&\textbf{43.3}&13.4&-&-&-\\
    SAE~\cite{kodirov2017semantic}&1.8&77.1&3.5&7.8&54.0&13.6&8.8&18.0&11.8&-&-&-\\
    \midrule
    Our RFF-GZSL&\textbf{22.0}&83.9&\textbf{34.8}&\textbf{26.2}&62.2&\textbf{36.9}&\textbf{22.3}&35.1&\textbf{27.3}&\textbf{24.4}&\textbf{71.1}&\textbf{36.3}\\
    \bottomrule
  \end{tabular}
}
\end{table*}

\subsection{Comparison with the state-of-the-art}
Table~\ref{table:comp_sota} shows the state-of-art results of GZSL, in
which we select thirteen results published in recent two years for
comparison. We organize the compared methods into two groups: (1) five 
non-feature generation methods and (2) eight feature generation based
methods. We compare our redundancy-free feature generation results
with these recent GZSL results. 

\begin{figure}[t]
 \centering
   \begin{subfigure}[b]{0.23\textwidth}
     \includegraphics[width=\linewidth]{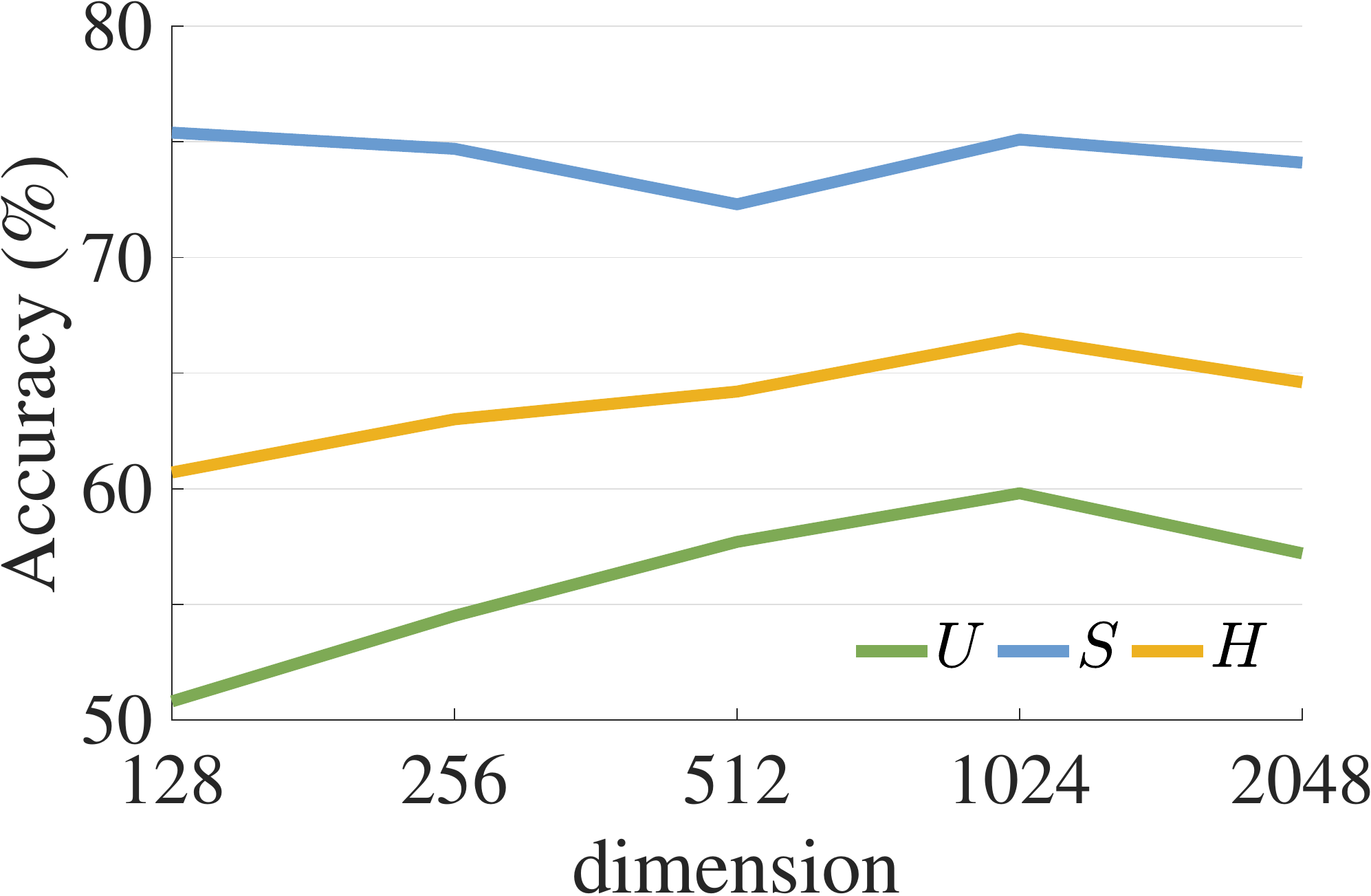}
     \caption{AWA}
     \label{fig:dim_awa}
   \end{subfigure}
   \begin{subfigure}[b]{0.23\textwidth}
     \includegraphics[width=\linewidth]{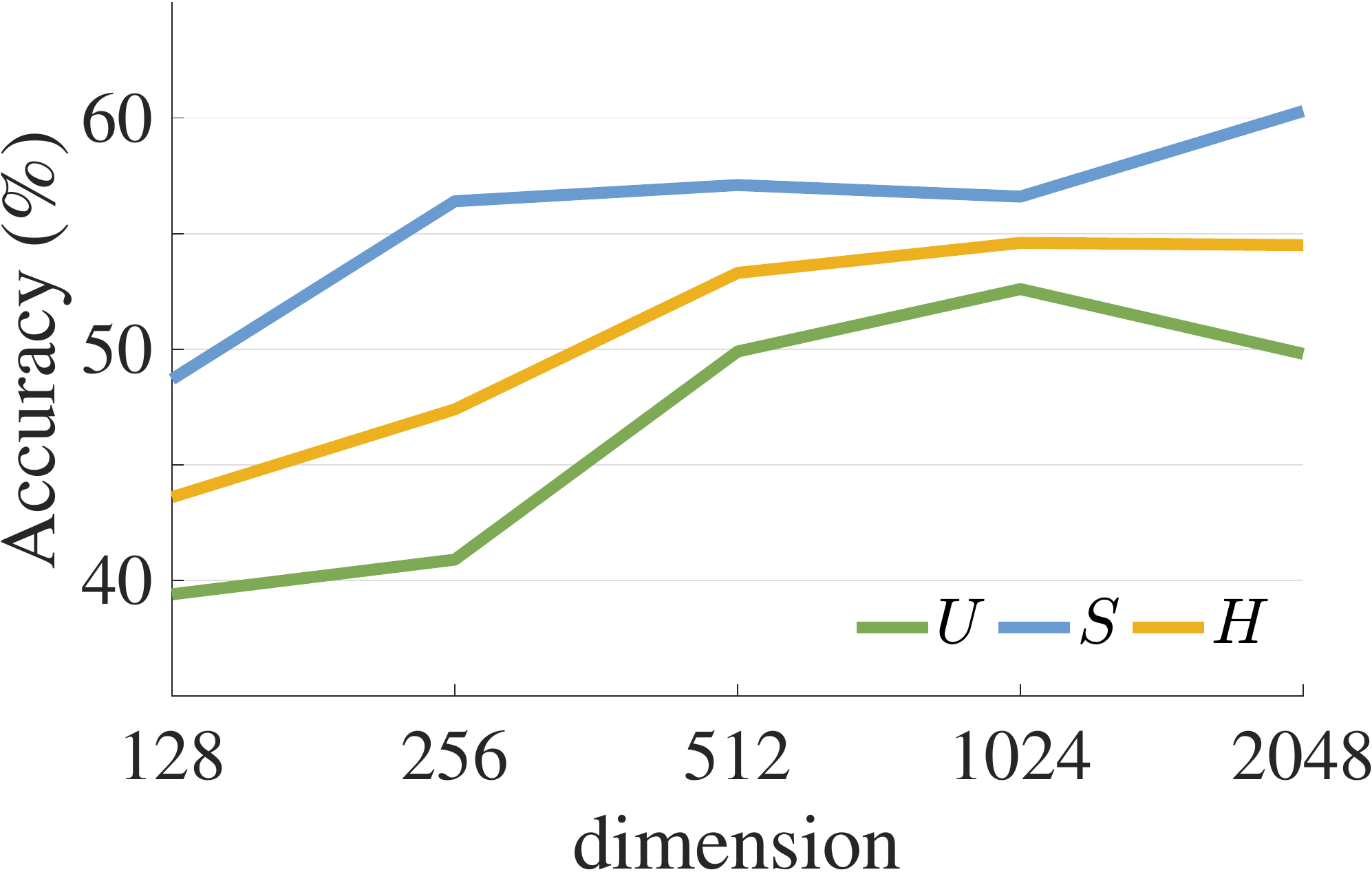}
     \caption{CUB}
     \label{fig:dim_cub}
   \end{subfigure}
   \begin{subfigure}[b]{0.23\textwidth}
    \includegraphics[width=\linewidth]{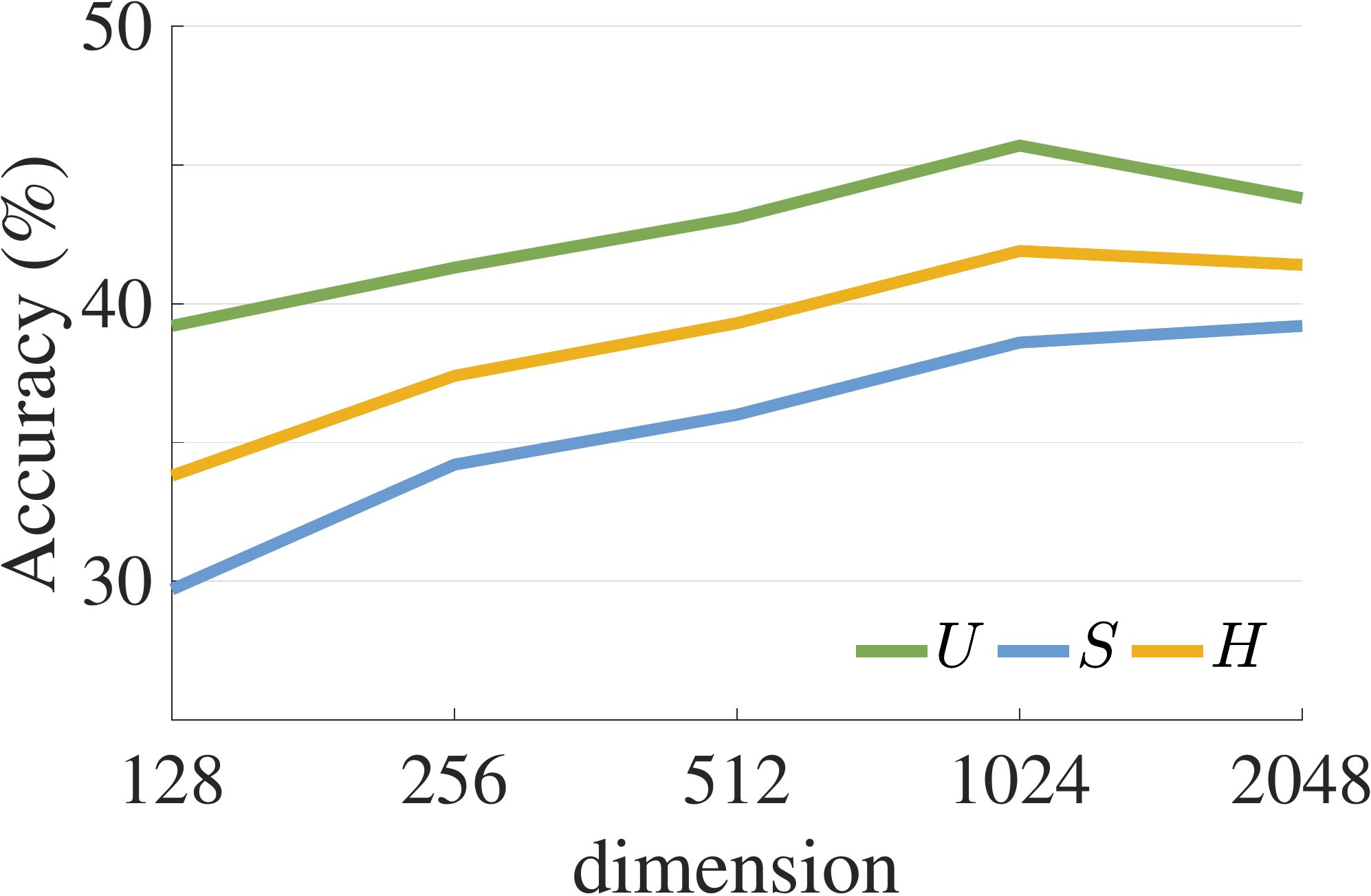}
    \caption{SUN}
    \label{fig:dim_sun}
  \end{subfigure}
  \begin{subfigure}[b]{0.23\textwidth}
    \includegraphics[width=\linewidth]{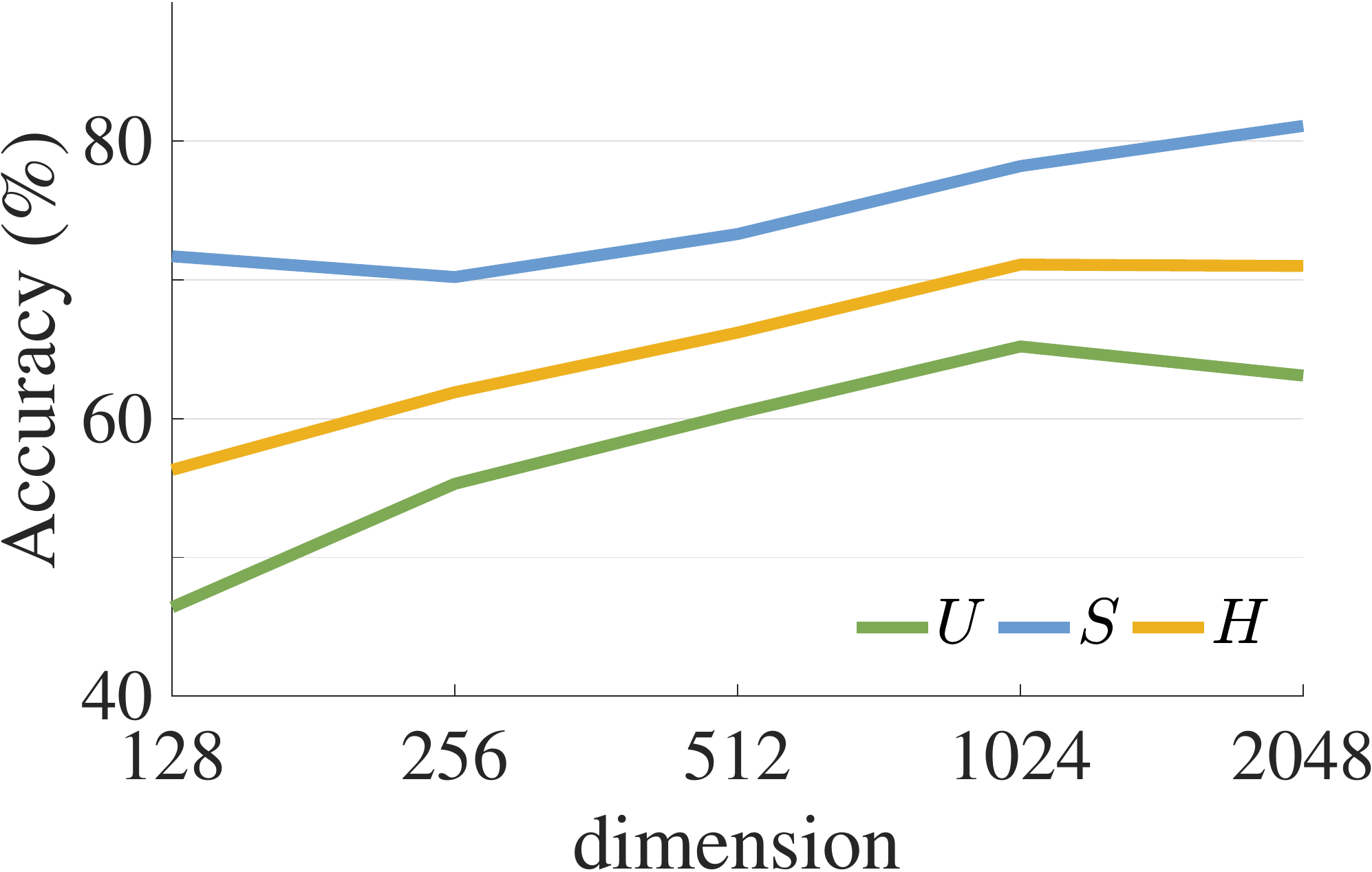}
    \caption{FLO}
    \label{fig:dim_flo}
  \end{subfigure}
 \caption{The influence of the redundancy-free feature dimensions on
   GZSL results, evaluated by our RFF-GZSL.}
 \vspace{-0.2cm}
 \label{fig:dimension_rff_space}
\end{figure}

We first compare our redundancy-free feature generation results with
the other feature generation methods. 
Our RFF-GZSL is
competitive compared with the feature generation
methods. Specifically, according to the harmonic mean results, our
RFF-GZSL can surpass the eight feature generation methods on
AWA, SUN and FLO. On CUB, our RFF-GZSL is only lower than
GMN~\cite{sariyildiz2019gradient} and
SABR~\cite{paul2019semantically}. 
And on FLO, our method outperform the second best method by a large margin.
Then, we compare our RFF-GZSL with
the non-feature generation methods. 
Our method can achieve the  best results evaluated on the unseen classes
($U$) and the harmonic mean ($H$) on CUB and SUN. 
And our method achieves the second best on $U$ and $S$ on AWA.
The results demonstrates the
effectiveness of our RFF-GZSL.

In Figure~\ref{fig:syn_num}, we report our RFF-GZSL results
under different numbers of synthesized samples per unseen class. 
When the amount of synthetic samples is small, the $U$ and $H$
results are low due to the data imbalance problem. As the number of
synthetic features increases, the $U$ and $H$ results improve
significantly, which means our redundancy-free feature generation
method can deal with the data imbalance problem in GZSL.

We also evaluate our feature generation method, RFF-GZSL, with
different dimensions of redundancy-free feature space, as shown in
Figure~\ref{fig:dimension_rff_space}. When the dimension is small, our
performances on four datasets are low. As the dimension increases,
the performances on these four datasets get better. 
With the dimension of the redundancy-free feature space
equal to 1,024, we can already achieve the satisfactory GZSL results
on the four datasets.

\begin{figure}[t]
  \centering
  \begin{subfigure}[b]{0.23\textwidth}
    \includegraphics[width=\linewidth]{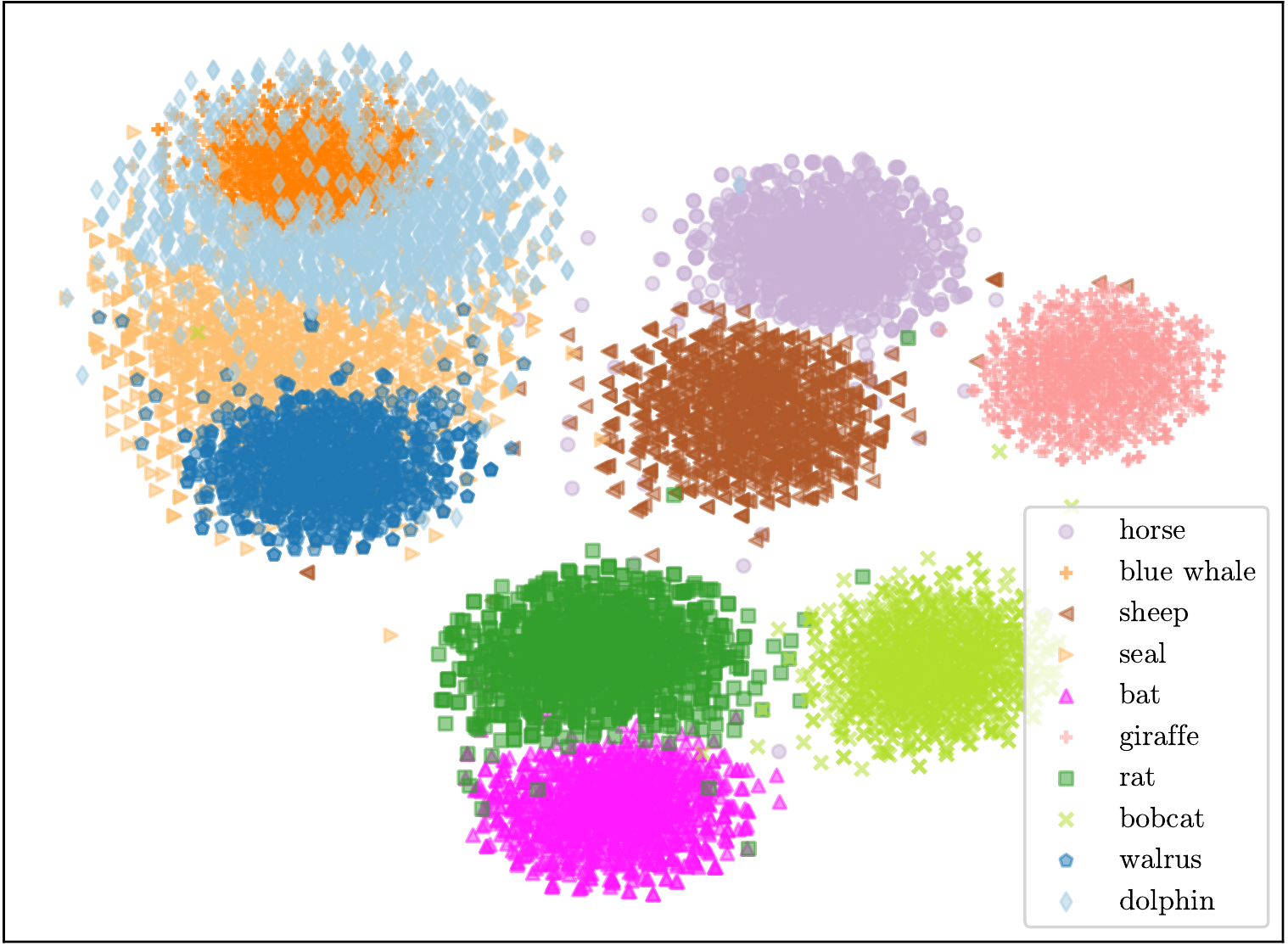}
    \caption{f-CLSWGAN}
    \label{fig:visual_baseline}
  \end{subfigure}
  \begin{subfigure}[b]{0.23\textwidth}
   \includegraphics[width=\linewidth]{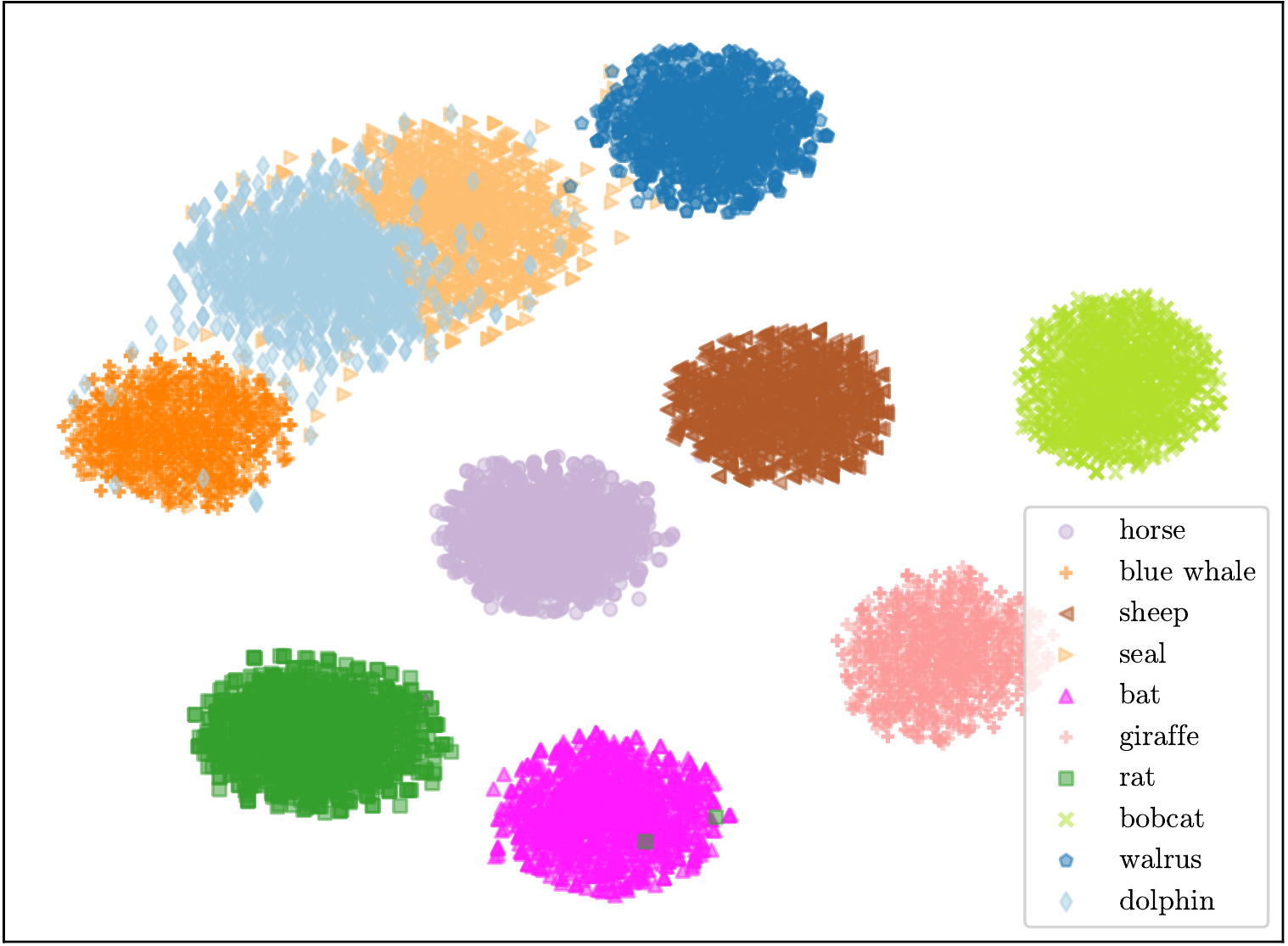}
   \caption{Our RFF-GZSL}
   \label{fig:visual_our_center}
 \end{subfigure}
  \caption{Visualization of the synthetic feature distributions of
    f-CLSWGAN~\cite{xian2018feature} and our RFF-GZSL on AWA.}
  \vspace{-0.2cm}
  \label{fig:visual_generated}
\end{figure}

\begin{figure*}[t]
  \centering
  \includegraphics[width=0.8\linewidth]{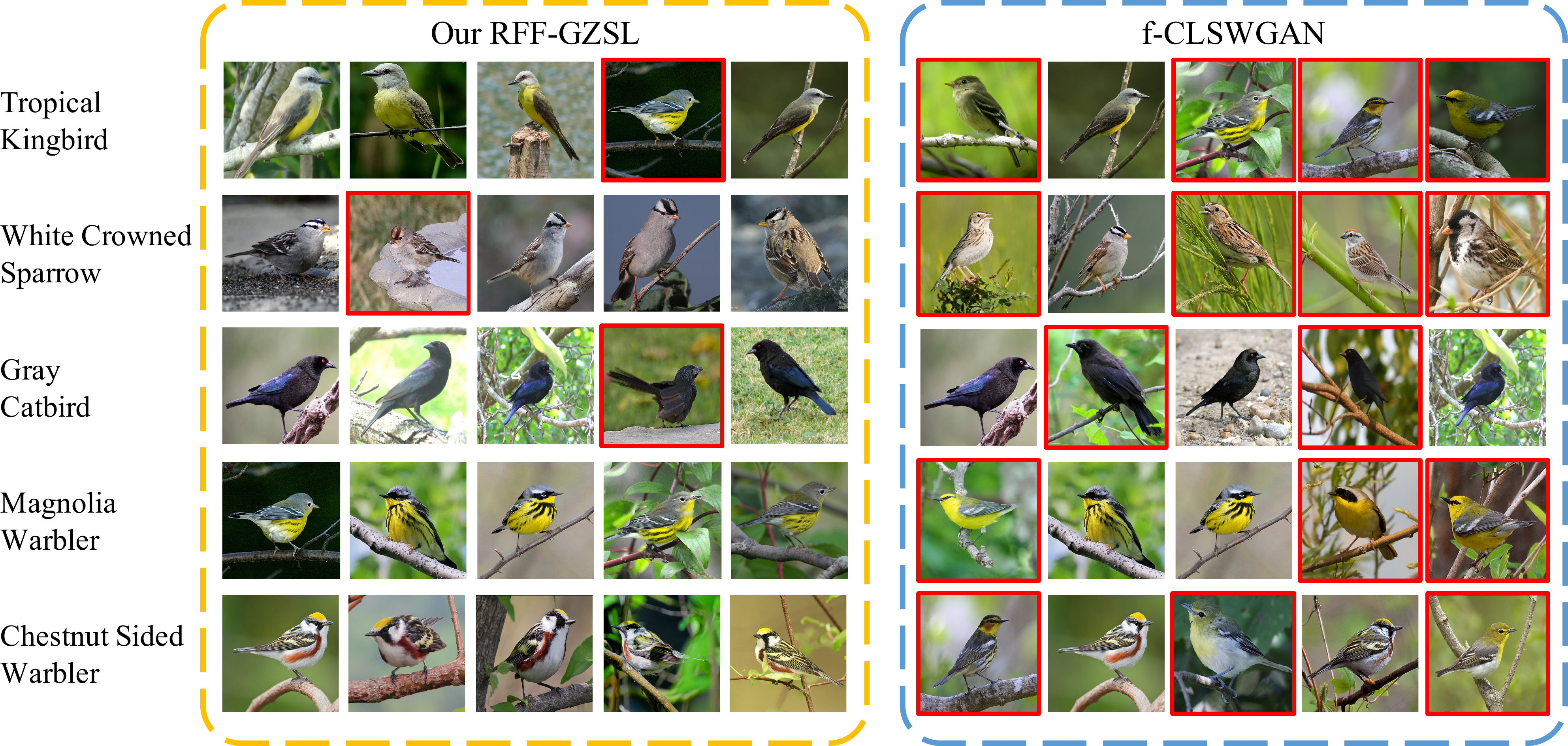}
  \caption{The results of image retrieval. Each row contains the Top-5
    retrieved images of a specific class. The image with a red box
    means a wrong retrieval.}
  \vspace{-0.2cm}
  \label{fig:retrieval}
\end{figure*}

\subsection{Comparison with traditional ZSL methods}
In this section, we compare the redundancy-free semantic embedding
model with several traditional ZSL methods in the new generalized ZSL
scenario. Table~\ref{tab:em_gzsl} shows the compared results. It can
be seen that the traditional ZSL methods usually perform poor in the
GZSL setting. Especially, all traditional ZSL methods in
Table~\ref{tab:em_gzsl} can achieve a high performance ($S$) on the
seen classes, but perform poor ($U$) on the unseen classes, resulting in a
low harmonic mean ($H$) for GZSL. Our redundancy-free semantic
embedding model can enhance the performance of the traditional
semantic embedding methods by reducing the redundancy information in
the original visual features. Specifically, our method is built upon
SJE~\cite{akata2015evaluation}; compared with
SJE~\cite{akata2015evaluation}, our redundancy-free semantic embedding
method can improve the GZSL results significantly on AWA and FLO with
almost 15\% enhancement. Finally, our method can surpass all compared
traditional ZSL methods on the new GZSL task.

\subsection{Visualization Results}
\paragraph{Feature visualization}
We visualize the features used in the final GZSL classification with
t-SNE~\cite{maaten2008visualizing}. We compare the visualization
result of our redundancy-free feature generation with
f-CLSWGAN~\cite{xian2018feature} on AWA to investigate the structure
of generated features. Concretely, for each unseen class on AWA, we
use the learned composite feature generator in our method to
synthesize 1,000 features in the redundancy-free feature space, while
we apply f-CLSWGAN to synthesize 1,000 features in the original visual
feature space. Since the dimension of the visual feature space is
2,048, for a fair comparison, we let our redundancy-free feature space
have the same  dimension. The results of f-CLSWGAN and ours are shown
in Figure~\ref{fig:visual_baseline} and
Figure~\ref{fig:visual_our_center}, respectively. As shown in
Figure~\ref{fig:visual_baseline}, in the generated visual feature
space of f-CLSWGAN, the feature distributions of four animal
categories, i.e. blue whale, seal, walrus and dolphin, are highly
overlapping. After reducing the redundancy information from the visual
features, as shown in~\ref{fig:visual_our_center}, these four and
other unseen categories can be easily separated in our redundancy-free
feature space.

\vspace{-0.2cm}
\paragraph{Image Retrieval}
We compare our RFF-GZSL with f-CLSWGAN~\cite{xian2018feature} on the
image retrieval task on CUB. Specifically, we use our RFF-GZSL to
synthesize 10 features in the redundancy-free feature space for each
unseen CUB class and then we apply the mean of these 10 features to
query the top-5 images, which have been mapped in the same
redundancy-free feature space. We do the same thing using
f-CLSWGAN~\cite{xian2018feature}, but this time the synthetic features 
locating in the original visual feature
space. Figure~\ref{fig:retrieval} shows the top-5 retrieval results 
of five bird example categories. The retrieval results of our method
are more accurate than f-CLSWGAN, demonstrating that the learned
redundancy-free features in our method are more discriminative than
the original visual features.

\section{Conclusion}
In this work, we have proposed to learn the redundancy-free features
for generalized zero-shot object recognition. We accomplish it by
learning a mapping function to map the original visual features to a
new redundancy-free feature space. We bound the statistical dependence
between these two feature spaces to remove the redundant information
from the visual features. Our method can integrate with existing GZSL
frameworks. The performance of conventional semantic embedding methods
has been promoted significantly using the redundancy-free
features. 
Our
redundancy-free feature generation model can achieve the
state-of-the-art or competitive results on four benchmarks. Also, the visualization
results, including the feature distribution and the image retrieval,
further demonstrate the effectiveness of learning the redundancy-free
features for GZSL.

\section*{Acknowledgment}
This work was supported by the National Science Fund of China under Grant Nos. U1713208, 61876085 and 61906092. It was also supported by China Postdoctoral Science Foundation (Grant No. 2017M621748 and 2019T120430).

{\small
\bibliographystyle{ieee_fullname}
\bibliography{egbib}
}
\end{document}